\documentclass[letterpaper]{article} % DO NOT CHANGE THIS
\usepackage{aaai2027}  % DO NOT CHANGE THIS
\usepackage[hyphens]{url}  % DO NOT CHANGE THIS
\usepackage{graphicx} % DO NOT CHANGE THIS
\usepackage{natbib}  % DO NOT CHANGE THIS AND DO NOT ADD ANY OPTIONS TO IT
\usepackage{caption} % DO NOT CHANGE THIS AND DO NOT ADD ANY OPTIONS TO IT
\usepackage{algorithm}
\usepackage{algorithmic}
\usepackage{amsmath,amssymb}
\usepackage{multirow}
\usepackage[table]{xcolor}
\usepackage{newfloat}
\usepackage{listings}
\DeclareCaptionStyle{ruled}{labelfont=normalfont,labelsep=colon,strut=off} % DO NOT CHANGE THIS
\floatstyle{ruled}
\newfloat{listing}{tb}{lst}{}
\floatname{listing}{Listing}

\usepackage{booktabs}

\newcommand{\blfootnote}[1]{%
  \begingroup
  \renewcommand{\thefootnote}{}%
  \footnotetext{#1}%
  \addtocounter{footnote}{-1}%
  \endgroup
}

\title{Pave-GRPO: Beyond Instantaneous Guidance through Principled Average Velocity Decomposition}

\author{
Pengyang Ling\textsuperscript{1}\textsuperscript{*},
Jiazi Bu\textsuperscript{2}\textsuperscript{*},
Yujie Zhou\textsuperscript{2}\textsuperscript{*},
Yibin Wang\textsuperscript{3},
Zhenyu Hu\textsuperscript{4},
Zihan Zhang\textsuperscript{5},
Yi Jin\textsuperscript{1},
Huaian Chen\textsuperscript{1}\textsuperscript{\textdagger},
Yuhang Zang\textsuperscript{6}\textsuperscript{\textdagger}
}

\affiliations{
\textsuperscript{1}University of Science and Technology of China,
\textsuperscript{2}Shanghai Jiao Tong University,
\textsuperscript{3}Fudan University,
\\
\textsuperscript{4}Harbin Institute of Technology,
\textsuperscript{5}Beihang University,
\textsuperscript{6}Shanghai AI Laboratory
}

\begin{document}

\twocolumn[{%
\renewcommand\twocolumn[1][]{#1}%
\maketitle

{\centering
\captionsetup{aboveskip=2pt}
\includegraphics[width=\textwidth]{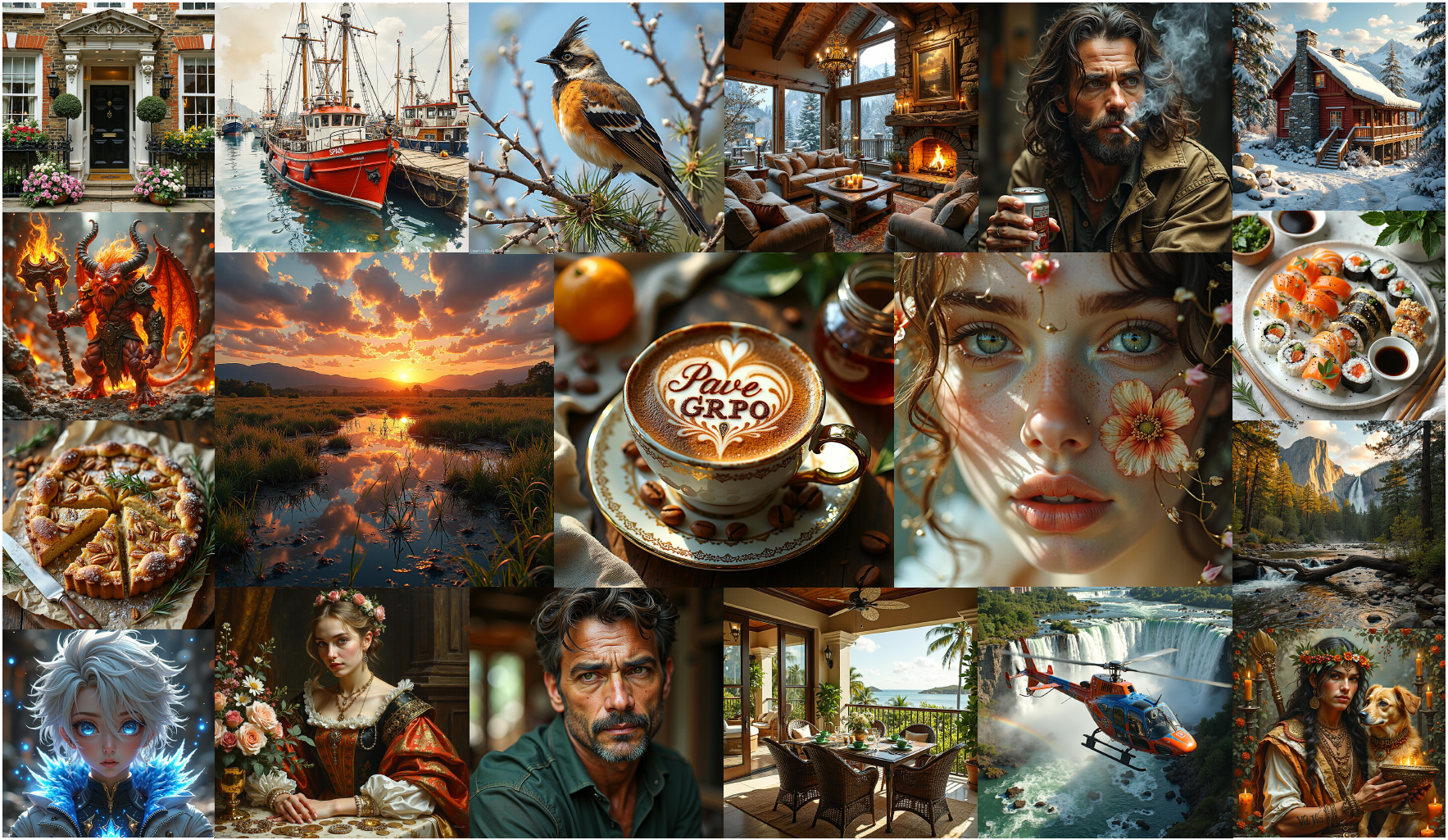}
\captionof{figure}{
Gallery of Pave-GRPO. Our Pave-GRPO post-training markedly improves T2I models
(FLUX.1-dev) on both global layout coherence and fine-grained detail fidelity.
All samples are generated with the same random seed (42), and the corresponding
prompts are provided in the supplementary material.
}
\label{fig:teaser}\par
\smallskip
}
}]

\blfootnote{%
\textsuperscript{*}Equal contribution.
\quad
\textsuperscript{\textdagger}Corresponding author.
}

\begin{abstract}
Post-training via Group Relative Policy Optimization (GRPO) has emerged as a powerful paradigm for aligning flow-based generative models with human preferences. However, the iterative denoising nature of flow models incurs substantial costs when generating group rollouts for policy-gradient updates, compelling existing methods to train with extremely few denoising steps. This temporal sparsity severely restricts preference optimization: reward feedback can only reach a handful of stages per trajectory, leaving the vast majority of intermediate denoising steps without direct supervision and thus compromising alignment granularity. To address this, we propose \textbf{Pave-GRPO}, which reformulates the GRPO objective through \textbf{p}rincipled \textbf{a}verage \textbf{v}elocity d\textbf{e}composition. Rather than generating expensive high-step rollouts, we maintain efficient few-step group sampling but decompose each coarse transition into an equivalent ensemble of finer sub-trajectories spanning multiple intermediate timesteps, propagating reward feedback to a denser set of temporal stages for more comprehensive preference alignment. Crucially, this incurs no additional stochastic rollout generation or reward evaluation: the decomposed sub-trajectories are constructed analytically around the observed transitions, requiring only a small number of extra velocity-network evaluations during the policy update. This design offers two benefits: (i) rollout-free horizon expansion: through the direct reuse of few-step group samples and their associated rewards, Pave-GRPO significantly broadens the effective optimization scope under a fixed sampling and reward budget; and (ii) comprehensive temporal supervision: by equivalently decomposing an instantaneous velocity target into a multi-timestep ensemble, it distributes reward signals across more intermediate stages of the denoising process, enabling finer-grained and more thorough preference optimization. Extensive experiments validate that Pave-GRPO effectively advances preference alignment across different reward settings, offering comprehensive performance enhancement.
\end{abstract}

\section{Introduction}
Recent advances in diffusion and flow models have enabled high-fidelity and flexible image synthesis across diverse domains. 
Correspondingly, the precise alignment of these generic pre-trained models with human preferences or task-specific constraints remains a non-trivial challenge. 
To this end, extensive research has explored post-training preference alignment techniques, which leverage feedback on model outputs to steer sampling trajectories toward desired endpoints. 
Such feedback can be derived from either direct human judgment~\cite{xu2023imagereward,liang2024rich} or automated ratings from reward models~\cite{lin2024evaluating,sun2025dreamsync}, both reflecting specific preferences regarding the generated content. 
To leverage outcome feedback for optimization, a straightforward approach is pairwise comparison, which increases the likelihood ratio of preferred samples over non-preferred ones, termed Direct Preference Optimization (DPO)~\cite{majumder2024tango,wallace2024diffusion,huang2025patchdpo}. However, such comparisons only provide ranking information, making it difficult to distinguish between marginally and substantially better samples, thereby limiting optimization granularity. To address this, Group Relative Policy Optimization (GRPO)~\cite{liu2025flow,li2025mixgrpo,he2025tempflow} samples a group of outputs from the same prompt and learns relative advantages through intra-group normalization. This eliminates shared baseline interference while reducing variance, thus improving both training stability and performance ceiling.

Specifically for flow models, existing approaches~\cite{xue2025dancegrpo,li2025branchgrpo,zhou2025g2rpo} typically convert the deterministic Ordinary Differential Equation (ODE) sampling into Stochastic Differential Equation (SDE) sampling to introduce the exploratory randomness required by GRPO. Nevertheless, due to the inherently multi-step iterative denoising process of flow models and GRPO's reliance on large group sizes for accurate intra-group advantage estimation, generating a rollout group remains computationally expensive. This compels existing methods to adopt significantly fewer training steps compared to inference (for instance, 16 training steps versus the default 50 inference steps for FLUX.1-dev). Consequently, each trajectory provides only a handful of supervised transitions, leaving the vast majority of intermediate denoising stages without direct reward guidance. Such temporal sparsity severely constrains fine-grained preference alignment, as the optimization signal is compressed into sparse velocity updates at isolated timesteps, rather than propagating across the fine-grained denoising trajectory.

In this work, we introduce \textbf{Pave-GRPO} to address this limitation. Rather than generating expensive high-step samples, we maintain efficient few-step group sampling but decompose each coarse transition into equivalently consecutive sub-fragments. The name Pave-GRPO reflects that: each coarse transition embodies an average velocity over its time interval, and our decomposition paves it into the finer instantaneous sub-velocities that compose it. This design stems from two observations. First, the iterative denoising process of a flow model is inherently a Markov process, in which each intermediate state serves as a sufficient statistic for the subsequent state, making these intermediate states reusable as shared targets for trajectory decomposition. Second, both pure SDE and hybrid ODE--SDE sampling yield tractable Gaussian probability paths. This allows the likelihood of few-step samples to be evaluated exactly under varying step configurations, enabling the recycling of advantages from cheap rollouts as supervision signals for multi-step optimization. The resulting decoupling yields two gains: (i) expanded effective optimization horizon without increased sampling cost, lifting the preference alignment ceiling; and (ii) comprehensive temporal supervision via shared intermediate regression targets, which distributes reward feedback across a denser set of temporal stages and enables finer-grained preference optimization. Extensive experiments confirm that Pave-GRPO consistently outperforms its competitors.

\section{Related work}
\subsection{Diffusion and flow models in visual generation}
Diffusion models~\cite{ho2020denoising,song2020denoising,song2020score,dhariwal2021diffusion,bu2025hiflow}
have become a dominant paradigm in generative modeling by learning to reverse an iterative noising process,
enabling high-fidelity synthesis across diverse modalities 
including images~\cite{rombach2022high,podell2023sdxl,flux2024} and video~\cite{guo2023animatediff,chen2024videocrafter2,wan2025wan,Zhou_2025_ICCV}. 
To mitigate the computational burden of operating in high-dimensional pixel spaces, 
Latent Diffusion Models (LDMs)~\cite{rombach2022high, podell2023sdxl} perform the denoising process within a compressed latent representation.
More recently, flow matching~\cite{esser2024scaling,lipman2022flow,liu2022flow} emerges as a compelling alternative framework,
which learns a continuous velocity field that transports probability mass from a noise distribution to the data distribution.
This paradigm has proven highly effective, achieving state-of-the-art performance in high-quality image generation, 
as evidenced by models such as Stable Diffusion 3~\cite{esser2024scaling}, FLUX series~\cite{flux2024, flux-2-2025}, 
and Qwen-Image~\cite{wu2025qwen}, as well as in large-scale video synthesis with models 
including HunyuanVideo~\cite{hunyuanvideo2025}, Wan series~\cite{wan2025wan} and LongCat-Video~\cite{team2025longcat}.

\subsection{Preference alignment in flow models}
Aligning diffusion and flow models with human preferences has evolved from early 
PPO-based policy gradients~\cite{black2023training,schulman2017proximal} 
and DPO variants~\cite{peng2025sudo,rafailov2023direct,wallace2024diffusion} toward more efficient online reinforcement learning frameworks, 
such as Group Relative Policy Optimization (GRPO)~\cite{shao2024deepseekmath}. 
To adapt GRPO for flow matching, foundational works like Flow-GRPO~\cite{liu2025flow}
and Dance-GRPO~\cite{xue2025dancegrpo} reformulate deterministic ODE sampling into equivalent SDE trajectories,
facilitating stochastic exploration while preserving marginal distributions. 
Subsequent efforts have refined this paradigm from multiple perspectives. 
For instance, Flow-CPS~\cite{wang2025coefficients} addresses the noise inconsistency in SDE sampling to prevent residual artifacts, 
while MixGRPO~\cite{li2025mixgrpo} introduces a hybrid ODE-SDE strategy to enhance training efficiency. 
Further improvements include tree-structured rollouts in BranchGRPO~\cite{li2025branchgrpo} for gradient concentration,
fine-grained credit assignment in TempFlow-GRPO~\cite{he2025tempflow} and Granular-GRPO~\cite{zhou2025g2rpo},
as well as dense multi-view reward mapping in MV-GRPO~\cite{bu2026sparse} without costly sample regeneration. 
Despite these advances, existing approaches employ few-step denoising during training to reduce cost, yet require high-step inference for quality. 
Hence, only a small number of transitions in each trajectory are supervised, leaving the rest of the intermediate denoising stages without direct reward signals.
This temporal sparsity intrinsically limits the granularity of preference alignment. 
To address this, our Pave-GRPO equivalently decomposes a low-step denoising trajectory into consecutive sub-fragments, decoupling the generation step budget from optimization granularity.

\section{Preliminaries}

\subsection{Sampling Formulations}

The standard denoising procedure in flow models advances from time step $t$ to $t-\Delta t$ (with $\Delta t > 0$) via a deterministic Ordinary Differential Equation (ODE), which can be expressed as:
\begin{equation}\label{eq:ode}
x_{t-\Delta t} = \mathcal{F}_{\text{ode}}(x_t, \Delta t) = x_t - \underbrace{v_\theta(x_t)\,\Delta t}_{\text{deterministic drift}},
\end{equation}
where $x_t$ and $x_{t-\Delta t}$ denote the latent states at $t$ and $t-\Delta t$, respectively, and $\mathcal{F}_{\text{ode}}(\cdot)$ denotes the ODE-based denoising operator driven by the predicted velocity field $v_\theta(x_t)$.
Since such deterministic sampling fails to support the exploratory mechanism for reinforcement learning, 
Flow-GRPO~\cite{liu2025flow} and Dance-GRPO~\cite{xue2025dancegrpo} 
utilize Stochastic Differential Equation (SDE) sampling
to inject randomness. 
Two widely used variants are the \emph{marginal-preserving} SDE, 
$\mathcal{F}_{\text{sde}}^{\text{mar}}(\cdot)$~\cite{liu2025flow}, and the \emph{SNR-preserving} SDE,
$\mathcal{F}_{\text{sde}}^{\text{snr}}(\cdot)$~\cite{wang2025coefficients}. 
The former maintains consistency with the marginal distribution of the ODE, while the latter preserves the signal-to-noise ratio. 
Mathematically, their denoising operators are formulated as:
\begin{equation}\label{eq:sde_md}
\begin{split}
&\mathcal{F}_{\text{sde}}^{\text{mar}}(x_t, \Delta t) 
= x_t - \Bigl[v_\theta(x_t) \\
&\quad + \frac{\sigma_t^2}{2t}\bigl(x_t + (1-t)v_\theta(x_t)\bigr)\Bigr]\Delta t 
+ \sigma_t\sqrt{\Delta t}\,\boldsymbol{\epsilon},
\end{split}
\end{equation}

\begin{equation}\label{eq:sde_snr}
\begin{split}
&\mathcal{F}_{\text{sde}}^{\text{snr}}(x_t, \Delta t) 
= (1-t+\Delta t)\bigl(x_t - t v_\theta(x_t)\bigr) \\
&\quad + \sqrt{(t{-}\Delta t)^2 {-} \sigma_t^2}\,\bigl(x_t {+} (1{-}t)v_\theta(x_t)\bigr) + \sigma_t\boldsymbol{\epsilon},
\end{split}
\end{equation}
where $\boldsymbol{\epsilon}\sim\mathcal{N}(\mathbf{0},\mathbf{I})$ is Gaussian noise enabling random exploration, and $\sigma_t$ governs the level of stochasticity at $t$.

\subsection{GRPO Reinforcement Learning}
Following Flow-GRPO~\cite{liu2025flow}, 
we formulate the flow matching process as a Markov Decision Process (MDP). Given a condition $c$, the generation trajectory is defined as $\Gamma = (s_T, a_T, \dots, s_0)$, where the state $s_t = (x_t, t, c)$ starts from a Gaussian prior $x_T \sim \mathcal{N}(\mathbf{0}, \mathbf{I})$. The action $a_t$ corresponds to the denoising step governed by the policy $\pi_\theta$. Unlike standard deterministic flow matching, we adopt the SDE-based sampling described in Eq.~\ref{eq:sde_md} to introduce necessary stochasticity for exploration.

In the GRPO framework, the policy is optimized using intra-group relative advantages, eliminating the need for a separate value network. For a given prompt $c$, we sample a group of $G$ trajectories $\{\Gamma^i\}_{i=1}^G$ using the current policy $\pi_{\theta_{\text{old}}}$. The advantage for the $i$-th sample is computed by normalizing the rewards within the group:
\begin{equation}
\hat{A}^i = \frac{R(\mathbf{x}_0^i, c) - \text{mean}(\{R(\mathbf{x}_0^j, c)\}_{j=1}^G)}{\text{std}(\{R(\mathbf{x}_0^j, c)\}_{j=1}^G)},
\end{equation}
where $R(\cdot)$ denotes the reward function evaluating the final generated sample $\mathbf{x}_0$.
The policy is then updated by maximizing the clipped surrogate objective with a KL penalty:
\begin{equation}
\begin{split}
&\mathcal{J}_{\text{GRPO}}(\theta) 
= \mathbb{E}_{c, \{\Gamma^i\}} \Biggl[ \frac{1}{G} \sum_{i=1}^G \frac{1}{T} \sum_{t=1}^T \biggl( \min\Bigl(r_t^i(\theta) \hat{A}^i, \\
&\quad \text{clip}\bigl(r_t^i(\theta), 1-\varepsilon, 1+\varepsilon\bigr) \hat{A}^i\Bigr) - \beta D_{\text{KL}}(\pi_\theta \| \pi_{\text{ref}}) \biggr) \Biggr].
\end{split}\label{eq:grpo_vanilla}
\end{equation}
where $r_t^i(\theta) = \frac{\pi_\theta(a_t^i|s_t^i)}{\pi_{\theta_{\text{old}}}(a_t^i|s_t^i)}$ is the importance sampling ratio, $\varepsilon$ is the clipping hyperparameter, and $\beta$ controls the divergence from the reference model $\pi_{\text{ref}}$.

\begin{figure}[ht]
  \centering
  \includegraphics[width=\linewidth]{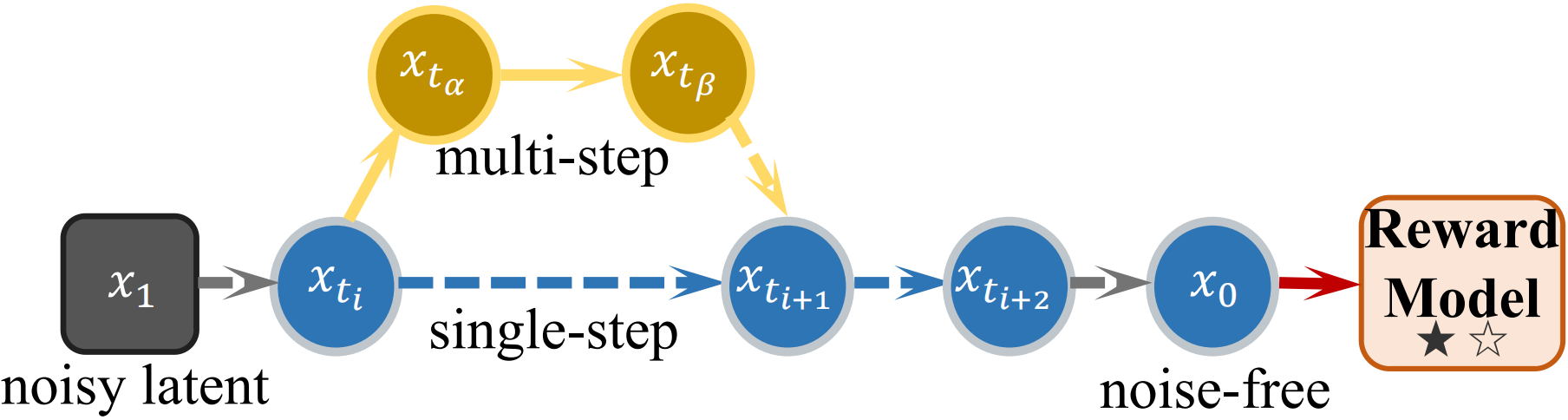}
  \caption{Iterative denoising process of flow models.}
  \label{fig:mdp}
\end{figure}

\begin{figure*}[t]
  \centering
  \includegraphics[width=\linewidth]{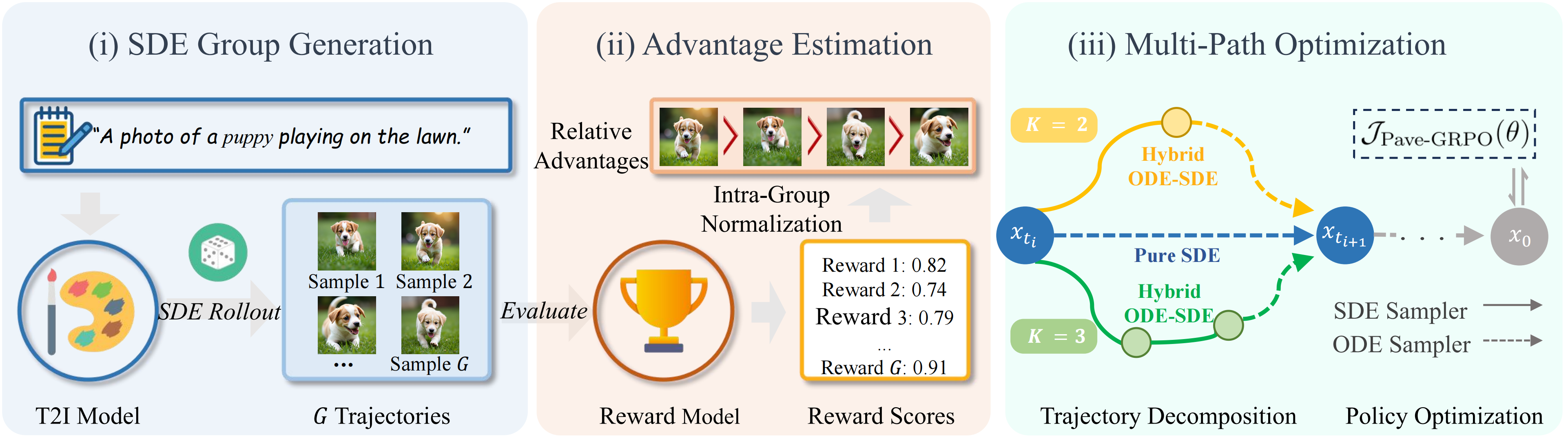}
  \caption{Overview of Pave-GRPO, which decomposes each pure-SDE transition into hybrid ODE--SDE sub-trajectories, propagating reward feedback to $k$ intermediate velocity predictions without extra sampling.}
  \label{fig:overview}
\end{figure*}

\section{Method}
\subsection{Reward Assessment in Markov Process}

As depicted by the blue arrows in Fig.~\ref{fig:mdp},
flow models generate content through an iterative denoising process, which can be formalized as a trajectory of discrete states $x_{t_i}$, indexed by time steps $t_i$ spanning [1,0].
Since most reward models are trained exclusively on noise-free images~\cite{xu2023imagereward,ma2025hpsv3,wang2026unified}, the reward score is usually computed at the endpoint of the denoising process $x_0$ and is subsequently back-propagated to the intermediate points along the trajectory.
Fundamentally, this iterative denoising process is a Markov Decision Process, which implies that: for any given intermediate state $x_{t_{i+1}}$, all future decisions and states depend entirely on the current state, independent of the historical trajectory:
\begin{equation}\label{eq:markov}
P(x_{t_{i+2}} \mid x_1, \dots, x_{t_{i+1}}) = P(x_{t_{i+2}} \mid x_{t_{i+1}}),
\end{equation}

Concretely, consider two alternative ways of reaching the same intermediate
state $x_{t_{i+1}}$ from $x_{t_i}$: (i) a single-step denoising $x_{t_i}
\rightarrow x_{t_{i+1}}$, and (ii) a multi-step denoising $x_{t_i}
\rightarrow x_{t_\alpha} \rightarrow x_{t_\beta} \rightarrow x_{t_{i+1}}$,
where $t_\alpha = \frac{2}{3}t_i + \frac{1}{3}t_{i+1}$ and $t_\beta =
\frac{1}{3}t_i + \frac{2}{3}t_{i+1}$. By the Markov property, the
continuation from $x_{t_{i+1}}$ to the endpoint $x_0$ is path-independent,
so both routes share the same future trajectory distribution and hence the
same endpoint reward. Consequently, $x_{t_{i+1}}$ can serve as a versatile
anchor: a single reward assessment of $x_0$ supervises every trajectory
passing through $x_{t_{i+1}}$, alleviating the high cost of repeated rollouts
from $x_{t_{i+1}}$ to $x_0$.

\subsection{Hybrid Sampling Trajectory}

Consider the denoising process from $t_i$ to $t_{i+1}$ with $\Delta t = t_i - t_{i+1}$. We investigate two alternative denoising trajectories connecting them:

\textbf{Pure SDE:} Run one-step SDE from $x_{t_i}$ to $x_{t_{i+1}}$, i.e.,
\begin{equation}\label{eq:pure}
x_{t_{i+1}}^{\text{pure}} = \mathcal{F}_{\text{sde}}^{(\cdot)}(x_{t_i}, \Delta t), \qquad (\cdot)\in\{\mathrm{mar},\,\mathrm{snr}\}.
\end{equation}

\textbf{Hybrid ODE--SDE:} Perform the ODE denoising operator for multiple steps, then take one final SDE operator to $t_{i+1}$ (operators are applied from left to right), i.e., 

\begin{equation}\label{eq:hybrid}
\begin{split}
&x_{t_{i+1}}^{\text{hybrid}} 
= \underbrace{\mathcal{F}_{\text{ode}}(x_{t_i}, \frac{\Delta t}{k}) \circ \cdots \circ\mathcal{F}_{\text{ode}}(x_{t_i - \frac{(k-2)\Delta t}{k}}, \frac{\Delta t}{k})}_{(k-1) \text{ step ODE}} \\
&\quad \circ \underbrace{\mathcal{F}_{\text{sde}}^{(\cdot)}(x_{t_i - \frac{(k-1)\Delta t}{k}}, \frac{\Delta t}{k})}_{\text{one step SDE}}, \quad (\cdot)\in\{\mathrm{mar},\,\mathrm{snr}\}.
\end{split}
\end{equation}

Our goal is to convert each sampled coarse transition into $k$ supervised sub-transitions \emph{without additional rollouts}. Intuitively, the single ODE step in Eq.~\ref{eq:ode} approximates the average velocity $\bar{v}_\theta = \frac{1}{\Delta t}\int_{t_{i+1}}^{t_i} v_\theta(x_t)\,dt$ over $[t_{i+1}, t_i]$ by the instantaneous velocity $v_\theta(x_{t_i})$ alone; the hybrid path instead estimates the same $\bar{v}_\theta$ with $k$ sub-velocities evaluated along the interval, decomposing each coarse average-velocity target into finer instantaneous constituents. Two properties of the above trajectories make this possible.

\paragraph{Closed-form Gaussian kernels.}
Since the ODE operator is deterministic and the SDE operator injects Gaussian noise, every transition in Eqs.~\ref{eq:pure}--\ref{eq:hybrid} is Gaussian: $x_{t_{i+1}}\sim\mathcal{N}(\mu_\theta^{(\cdot)},\Sigma^{(\cdot)})$. For SDE operator in Eq.~\ref{eq:sde_md}, denote $A_\theta(x_t)=v_\theta(x_t)+\frac{\sigma_t^2}{2t}\big(x_t+(1-t)v_\theta(x_t)\big)$, each step reads $x_{t-\Delta t}=x_t-A_\theta(x_t)\,\Delta t+\sigma_t\sqrt{\Delta t}\,\epsilon$, giving means as:

\begin{equation}\label{eq:means}
\mu_\theta^{\mathrm{pure}} \!=\! x_{t_i} \!-\! A_\theta(x_{t_i})\,\Delta t, \;
\mu_\theta^{\mathrm{hybrid}} \!=\! x_{t_m} \!-\! A_\theta(x_{t_m})\,\tfrac{\Delta t}{k},
\end{equation}
and parameter-free isotropic covariances $\Sigma^{\mathrm{pure}}=\sigma_{t_i}^{2}\Delta t\, I$ and $\Sigma^{\mathrm{hybrid}}=\sigma_{t_m}^{2}\frac{\Delta t}{k}\, I$. Here $x_{t_m}$ at $t_m=t_i-\frac{(k-1)\Delta t}{k}$ is produced by $k{-}1$ deterministic ODE sub-steps ($x_{s_{j+1}}=x_{s_j}-v_\theta(x_{s_j})\frac{\Delta t}{k}$ with $s_j=t_i-\frac{j\Delta t}{k}$, $j=0,\dots,k{-}2$). Hence $\mu_\theta^{\mathrm{hybrid}}$ depends on all $k$ velocity predictions $\{v_\theta(x_{s_j})\}_{j=0}^{k-1}$: the first $k{-}1$ generate $x_{t_m}$, and the last enters $A_\theta(x_{t_m})$. All intermediate states remain differentiable, allowing the hybrid log-likelihood to supervise all $k$ predictions jointly.
The $\mathcal{F}_{\text{snr}}$ in Eq.~\ref{eq:sde_snr} admits analogous Gaussian kernels, with means given by the linear coefficients and isotropic covariances.

\begin{figure*}[t]
  \centering
  \includegraphics[width=\linewidth]{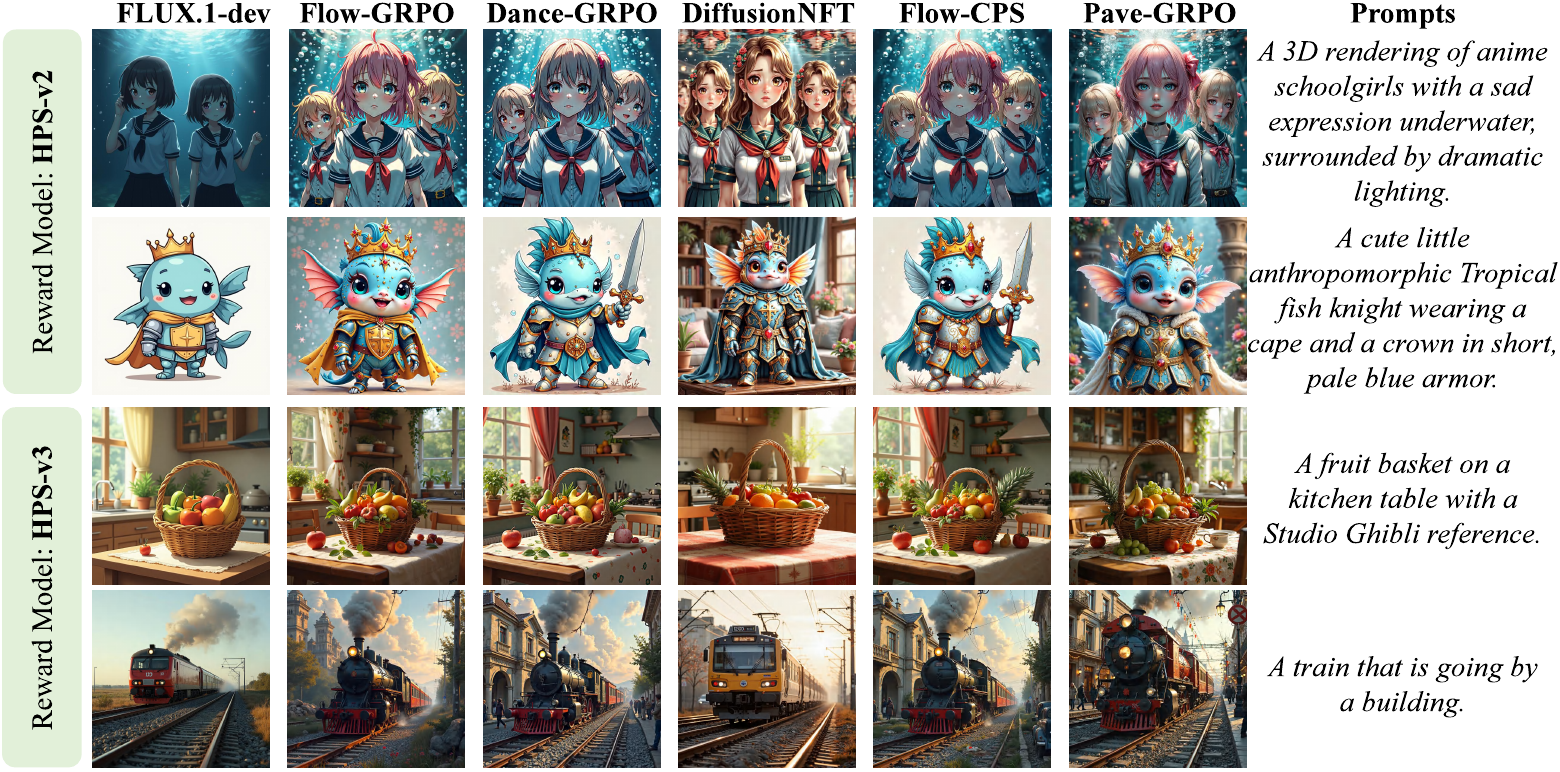}
  \caption{Visual comparison with competing methods. Pave-GRPO exhibits superior structural integrity and finer details.}
  \label{fig:visual_comparison}
\end{figure*}

\paragraph{Path-consistent endpoint statistics.}
Both SDE variants are agnostic to the denoising path at the endpoint: the marginal-preserving SDE shares its Fokker--Planck equation~\cite{oksendal2003stochastic} with the probability-flow ODE, while the SNR-preserving SDE fixes the endpoint SNR through its diffusion coefficients~\cite{wang2025coefficients}. Pure and hybrid paths from the same marginal at $t_i$ thus reach identical marginal/SNR at $t_{i+1}$, i.e., a hybrid sub-trajectory is a genuine sampling path of the model rather than an artificial construct.
The Markov property makes the endpoint reward independent of the path producing $x_{t_{i+1}}$, so a single reward assessment covers the whole transition. With the closed-form kernels, the likelihood of an already-sampled endpoint under any hybrid decomposition is then evaluable as follows:
\begin{equation}\label{eq:density}
p_\theta^{(k,\mathrm{hybrid})}(x_{t_{i+1}}^{\mathrm{pure}} \mid x_{t_i}) = \mathcal{N}(x_{t_{i+1}}^{\mathrm{pure}}; \mu_\theta^{\mathrm{hybrid}}, \Sigma^{\mathrm{hybrid}})
\end{equation}

\subsection{Training Pipeline}

As illustrated in Fig.~\ref{fig:overview}, Pave-GRPO operates in three stages:
(i)~\textbf{SDE Group Generation.}
For each prompt, we sample $G$ trajectories via few-step pure-SDE sampling, preserving GRPO's exploratory stochasticity at minimal cost.
(ii)~\textbf{Advantage Estimation.}
A reward model scores each terminal output; the rewards $\{r_g\}_{g=1}^G$ are group-normalized into relative advantages $\{\hat{A}^g\}_{g=1}^G$.
(iii)~\textbf{Multi-Path Optimization.}
Each coarse transition $x_{t_i}\!\to\!x_{t_{i+1}}$ serves as a fixed boundary condition for constructing hybrid ODE--SDE sub-trajectories with granularities $k\!\in\!\mathcal{S}$ (e.g., $\{2,3\}$); the closed-form Gaussian kernels give the exact likelihood of the observed endpoint $x_{t_{i+1}}$ under every decomposition without additional rollouts. The policy then increases the likelihood of high-advantage transitions along all decomposed paths, propagating reward feedback to the $k$ intermediate velocity predictions for dense temporal supervision.

For the \textbf{training objective}, Pave-GRPO augments the standard clipped surrogate objective with a hybrid decomposition term. Let $\mathcal{S}$ be the set of decomposition factors (e.g., $\{2,3\}$) and define $\mathrm{Surr}(r,A) = \min(rA, \mathrm{clip}(r, 1-\varepsilon, 1+\varepsilon)A)$. The overall objective is modeled as:
\begin{equation}
\begin{split}
& \mathcal{J}_{\text{Pave-GRPO}}(\theta) = \mathbb{E} \Biggl[ \frac{1}{GT} \sum_{i=1}^G \sum_{t=1}^T \biggl( \mathrm{Surr}(\rho_t^i(\theta), \hat{A}^i) \\
& \hspace{3.5em} + \frac{1}{|\mathcal{S}|} \sum_{k \in \mathcal{S}} \mathrm{Surr}(\eta_{t,k}^i(\theta), \hat{A}^i) - \beta \mathcal{D}_{\mathrm{KL}} \biggr) \Biggr],
\end{split}
\end{equation}
where $\rho_t^i(\theta)$ and $\eta_{t,k}^i(\theta)$ are the importance ratios of the pure and hybrid paths, respectively. Both ratios are evaluated at the same endpoint $x_{t-\Delta t}^i$ sampled from the pure-SDE rollout ($x_t^i$ and $x_{t-\Delta t}^i$ correspond to $x_{t_i}$ and $x_{t_{i+1}}$ in Eqs.~\ref{eq:pure}--\ref{eq:density}), which can be expressed as: 
\begin{equation}
\begin{aligned}
\rho_t^i(\theta) &= \frac{p_{\theta}(x_{t-\Delta t}^i \mid x_{t}^i)}{p_{\theta_{\mathrm{old}}}(x_{t-\Delta t}^i \mid x_{t}^i)}, \\
\eta_{t,k}^i(\theta) &= \frac{p_{\theta}^{(k,\mathrm{hybrid})}(x_{t-\Delta t}^i \mid x_{t}^i)}{p_{\theta_{\mathrm{old}}}^{(k,\mathrm{hybrid})}(x_{t-\Delta t}^i \mid x_{t}^i)},
\end{aligned}
\end{equation}
with the hybrid kernel $p^{(k,\mathrm{hybrid})}_\theta$ given in closed form by Eq.~\ref{eq:density}. The shared $\theta$ emphasizes that both kernels are parameterized by the same model and differ only in the denoising path: $\rho$ supervises the coarse-grained transition, while $\eta$ transfers the supervision into $k$ finer-grained sub-velocities through the hybrid trajectory decomposition.
% reaching intermediate velocities that would otherwise receive no optimization signal.

\begin{table*}[t]
\centering
\small
\setlength{\tabcolsep}{4.5pt}
\resizebox{0.95\textwidth}{!}{%
\begin{tabular}{@{}llccccccc@{}}
\toprule
Reward Model & Method & HPS-v2$\uparrow$ & HPS-v3$\uparrow$ & CLIP$\uparrow$ & ImageReward$\uparrow$ & UnifiedReward-v1$\uparrow$ & UnifiedReward-v2$\uparrow$ & Pick-Score$\uparrow$ \\
\midrule
None & FLUX.1-dev & 30.65 & 13.28 & 39.01 & 1.0546 & 3.6006 & 3.3810 & 22.68 \\
\midrule
\multirow{6}{*}{HPS-v2} 
& Flow-GRPO & 34.63 & 14.44 & \textbf{37.54} & 1.2435 & 3.5167 & 3.3975 & 22.84 \\
& Dance-GRPO & 34.30 & 14.47 & 37.24 & 1.2178 & 3.5129 & 3.3878 & 23.01 \\
& DiffusionNFT & 36.04 & 14.48 & 35.89 & 1.2859 & 3.4989 & 3.4064 & 22.61 \\
& Flow-CPS & 34.92 & 14.80 & 37.45 & 1.2561 & 3.5276 & 3.4242 & 23.11 \\
& Pave-GRPO ($\mathcal{F}_{\text{sde}}^{\text{mar}}$) & \textbf{36.40} & \textbf{15.12} & 36.72 & \textbf{1.3331} & 3.5396 & 3.4181 & \textbf{23.25} \\
& Pave-GRPO ($\mathcal{F}_{\text{sde}}^{\text{snr}}$) & 36.24 & 15.10 & 36.88 & 1.3003 & \textbf{3.5467} & \textbf{3.4362} & 23.14 \\
\midrule
\multirow{6}{*}{HPS-v3} 
& Flow-GRPO & 32.72 & 14.71 & 37.87 & 1.1475 & \textbf{3.5821} & 3.3955 & 23.07 \\
& Dance-GRPO & 32.75 & 14.67 & \textbf{38.13} & 1.1431 & 3.5460 & 3.3879 & 23.00 \\
& DiffusionNFT & 33.10 & 14.91 & 35.48 & 1.1300 & 3.4048 & \textbf{3.4133} & 22.64 \\
& Flow-CPS & 32.91 & 14.76 & 37.83 & 1.1598 & 3.5375 & 3.3883 & 23.02 \\
& Pave-GRPO ($\mathcal{F}_{\text{sde}}^{\text{mar}}$) & \textbf{33.44} & \textbf{15.18} & 37.22 & \textbf{1.1658} & 3.5431 & 3.3890 & \textbf{23.14} \\
& Pave-GRPO ($\mathcal{F}_{\text{sde}}^{\text{snr}}$) & 33.06 & 15.05 & 37.26 & 1.1421 & 3.4896 & 3.3705 & 22.85 \\
\midrule
\multirow{6}{*}{HPS-v2 \!+\! CLIP} 
& Flow-GRPO & 33.54 & 14.27 & 39.22 & 1.2496 & 3.6181 & 3.3919 & 23.03 \\
& Dance-GRPO & 33.18 & 14.18 & 39.32 & 1.2307 & 3.5848 & 3.3699 & 22.99 \\
& DiffusionNFT & 34.46 & 14.35 & 39.15 & 1.3080 & 3.6536 & 3.3720 & 23.10 \\
& Flow-CPS & 33.64 & 14.28 & 39.37 & 1.2889 & 3.6482 & 3.3877 & 23.14 \\
& Pave-GRPO ($\mathcal{F}_{\text{sde}}^{\text{mar}}$) & \textbf{34.84} & \textbf{14.53} & \textbf{39.57} & \textbf{1.3841} & \textbf{3.7020} & \textbf{3.3978} & \textbf{23.27} \\
& Pave-GRPO ($\mathcal{F}_{\text{sde}}^{\text{snr}}$) & 34.36 & 14.45 & 39.10 & 1.3115 & 3.6668 & 3.3955 & 23.15 \\
\midrule
\multirow{6}{*}{HPS-v3 \!+\! CLIP} 
& Flow-GRPO & 32.50 & 14.25 & 39.38 & 1.1988 & 3.6260 & 3.3774 & 23.02 \\
& Dance-GRPO & 31.89 & 13.86 & 39.14 & 1.1406 & 3.5859 & 3.3759 & 22.73 \\
& DiffusionNFT & 31.59 & 14.02 & 38.36 & 1.0552 & 3.5801 & 3.3801 & 22.54 \\
& Flow-CPS & 32.52 & 14.32 & 39.26 & 1.2222 & 3.6339 & 3.3715 & 23.13 \\
& Pave-GRPO ($\mathcal{F}_{\text{sde}}^{\text{mar}}$) & \textbf{33.15} & \textbf{14.62} & \textbf{39.58} & \textbf{1.2749} & \textbf{3.6591} & \textbf{3.3834} & \textbf{23.19} \\
& Pave-GRPO ($\mathcal{F}_{\text{sde}}^{\text{snr}}$) & 32.74 & 14.37 & 39.46 & 1.2318 & 3.6123 & 3.3741 & 23.08 \\
\bottomrule
\end{tabular}%
}
\caption{Quantitative comparison across four different reward settings.}
\label{tab:main_results}
\end{table*}

\begin{table*}[t]
\centering
\small
\setlength{\tabcolsep}{5pt}
\begin{tabular}{@{}l c cccccccccc@{}}
\toprule
Method & Overall$\uparrow$ & Style$\uparrow$ & Know.$\uparrow$ & Attri.$\uparrow$ & Act.$\uparrow$ & Rel.$\uparrow$ & Comp.$\uparrow$ & Gram.$\uparrow$ & Reason.$\uparrow$ & Layout$\uparrow$ & Text$\uparrow$ \\
\midrule
FLUX.1-dev & 59.86 & 84.60 & 85.76 & 64.32 & 61.22 & 66.37 & 46.26 & 58.69 & 27.06 & 70.71 & 33.62 \\
\midrule
Flow-GRPO & 62.65 & 85.80 & 89.08 & 70.19 & 66.06 & 69.67 & 51.03 & 60.70 & 29.36 & \textbf{73.88} & 30.75 \\
Dance-GRPO & 61.15 & 85.00 & 88.29 & 68.48 & 61.12 & 70.05 & 48.20 & 59.36 & 28.90 & 71.64 & 30.46 \\
DiffusionNFT & 59.24 & 74.20 & 89.72 & 67.41 & 62.83 & 65.10 & 44.33 & 57.35 & 27.29 & 72.57 & 31.61 \\
Flow-CPS & 62.65 & 86.10 & 90.03 & 69.66 & 65.87 & 71.32 & 52.19 & \textbf{61.36} & 29.13 & 73.51 & 27.30 \\
Pave-GRPO ($\mathcal{F}_{\text{sde}}^{\text{mar}}$) & \textbf{63.96} & 85.40 & \textbf{90.66} & \textbf{72.22} & \textbf{67.11} & \textbf{71.95} & \textbf{52.84} & 60.56 & 30.28 & 73.51 & \textbf{35.06} \\
Pave-GRPO ($\mathcal{F}_{\text{sde}}^{\text{snr}}$) & 63.23 & \textbf{86.40} & 89.87 & 70.51 & 65.68 & 71.45 & 50.26 & 60.56 & \textbf{33.26} & \textbf{73.88} & 30.46 \\
\bottomrule
\end{tabular}
\caption{Quantitative comparison on UniGenBench. All methods are trained with reward models: HPS-v3 + CLIP.}
\label{tab:unigenbench}
\end{table*}

\section{Experiments}
\subsection{Implementation Details}\label{subsec:implementation_details}

Following prior works~\cite{xue2025dancegrpo, zhou2025g2rpo}, we adopt the HPD dataset~\cite{wu2023human} (over 100K prompts) for training, with a held-out set of 400 prompts for testing, and additionally evaluate on UniGenBench~\cite{Pref-GRPO&UniGenBench}, which provides fine-grained assessment over 600 prompts across 10 dimensions. The experiments are conducted based on FLUX.1-dev~\cite{flux2024} and SD3.5-Medium~\cite{esser2024scaling}. Baselines include Flow-GRPO~\cite{liu2025flow}, Dance-GRPO~\cite{xue2025dancegrpo}, DiffusionNFT~\cite{zheng2025diffusionnft}, and Flow-CPS~\cite{wang2025coefficients}. 
For the evaluation metrics,
we report three groups: (i) CLIP~\cite{radford2021learning} for image--text alignment; (ii) reward-model scores, including ImageReward~\cite{xu2023imagereward}, HPS-v2/v3~\cite{wu2023human,ma2025hpsv3}, UnifiedReward-v1/v2~\cite{wang2025unified}, and PickScore~\cite{kirstain2023pick}; and (iii) UniGenBench~\cite{Pref-GRPO&UniGenBench}, where an MLLM assessor scores 10 dimensions (attribute binding, spatial layout, text rendering, etc.) for a holistic view of semantic fidelity and visual quality. We perform full-parameter training in \texttt{bfloat16} mixed precision with batch size 8, using AdamW with a learning rate of $2\times10^{-6}$. Pave-GRPO builds on the step-wise reward variant of Flow-GRPO for reliable credit assignment, with a fixed group size of 12 in rollouts. Following previous work~\cite{xue2025dancegrpo}, we use 16 denoising steps with pure-SDE sampling at timesteps $\{0,2,4,6\}$ during training, while evaluation uses 50 steps for higher quality. The noise intensity is $\sigma_t = 0.7\sqrt{\frac{t}{1-t}}$ for the \emph{marginal-preserving} SDE and $\sigma_t = (t-\Delta t)\sin(\frac{0.8\pi}{2})$ for the \emph{SNR-preserving} one. All baselines share identical training configurations.

\subsection{Qualitative Results}
The visual comparison in Fig.~\ref{fig:visual_comparison} shows that Pave-GRPO consistently outperforms competing methods in both visual fidelity and aesthetic appeal across different reward settings.
Under HPS-v2, Pave-GRPO renders the underwater anime schoolgirls with sharper facial features and richer details, while FLUX.1-dev produces comparatively dim and flat results. The tropical fish knight also shows richer material realism, with metallic reflections, fabric folds, and ambient lighting that Flow-GRPO and Dance-GRPO largely fail to reproduce.
This advantage also extends to results under HPS-v3. The proposed Pave-GRPO synthesizes the fruit basket containing more varied produce with nuanced surface textures and detailed kitchen surroundings, achieving stronger photorealism than the baselines. The generated train scene likewise presents a structurally coherent locomotive with improved environmental lighting and atmospheric depth.
These results confirm that Pave-GRPO effectively improves both subject fidelity and scene aesthetics.

\subsection{Quantitative Evaluation}
Tab.~\ref{tab:main_results} compares the proposed method under four reward settings, in which Pave-GRPO consistently outperforms competitors in the vast majority of metrics, validating its robustness and superiority. It achieves strong results under both \emph{marginal-preserving} and \emph{SNR-preserving} SDE frameworks, since the trajectory decomposition is agnostic to specific path constructions and applies uniformly to different sampling forms, thus providing excellent adaptability.
On UniGenBench, Pave-GRPO($\mathcal{F}_{\text{sde}}^{\text{mar}}$) achieves the highest overall score (63.96), which significantly surpasses the FLUX.1-dev baseline in Knowledge, Attribute, Action, and Text, and also exceeds Flow-GRPO by +1.31 overall. These performance gains demonstrate that the fine-grained optimization enhances complex semantic understanding and compositional generation beyond superficial aesthetic metrics. Furthermore,
for a supplementary cross-architecture validation, Tab.~\ref{tab:sd35} presents the results on SD3.5-Medium, in which the proposed Pave-GRPO gains the best scores on different rewards, confirming that the trajectory decomposition remains effective across different model architectures.

\begin{table}[t]
\centering
\begin{tabular}{@{}lccc@{}}
\toprule
Method & HPS-v3$\uparrow$ & CLIP$\uparrow$ & ImageReward$\uparrow$ \\
\midrule
SD3.5-Medium & 10.92 & 40.40 & 1.0049 \\
Flow-GRPO & 12.92 & 40.52 & 1.3288 \\
Dance-GRPO & 12.35 & 40.24 & 1.2313 \\
DiffusionNFT & 13.04 & 40.15 & 1.2706 \\
Flow-CPS & 12.98 & 40.24 & 1.3072 \\
Pave-GRPO ($\mathcal{F}_{\text{sde}}^{\text{mar}}$) & \textbf{13.48} & 40.60 & 1.3470 \\
Pave-GRPO ($\mathcal{F}_{\text{sde}}^{\text{snr}}$) & 13.46 & \textbf{40.63} & \textbf{1.3509} \\
\bottomrule
\end{tabular}
\caption{Quantitative comparison on SD3.5-Medium. All methods are trained with 16 steps using HPS-v3 + CLIP rewards and evaluated with the default 40 steps.}
\label{tab:sd35}
\end{table}

\begin{table}[t]
\centering
\begin{tabular}{@{}lccc@{}}
\toprule
Settings & HPS-v3$\uparrow$ & CLIP$\uparrow$ & Wall-clock time$\downarrow$ \\
\midrule
Flow-GRPO & 14.25 & 39.38 & 71s \\
FlowGRPO-T & 14.37 & 39.55 & 136s \\
FlowGRPO-G & 14.30 & 39.40 & 136s \\
$\mathcal{S}=\{2\}$ & 14.58 & 39.64 & 73s \\
$\mathcal{S}=\{2,3\}$ & 14.62 & 39.58 & 77s \\
$\mathcal{S}=\{2,3,4\}$ & 14.50 & 39.67 & 82s \\
$\mathcal{S}=\{2,3,4,5\}$ & 14.58 & 39.62 & 89s \\
\bottomrule
\end{tabular}
\caption{Ablations on $\mathcal{S}$. Reward setting: HPS-v3 + CLIP.}
\label{tab:ablations}
\end{table}

\subsection{Analysis and Ablations}

Pave-GRPO shares the same backbone and the same \emph{marginal-preserving} SDE with Flow-GRPO, differing only in the multi-path decomposition, thus the comparison in Tabs.~\ref{tab:main_results}--~\ref{tab:unigenbench} constitutes a direct ablation, in which Pave-GRPO provides stable performance improvement.
Tab.~\ref{tab:ablations} further shows that such gains are achieved economically: doubling the group size (FlowGRPO-G) or the denoising steps (FlowGRPO-T) nearly doubles the per-step cost ($136$s vs.\ $71$s) yet improves HPS-v3 by at most $0.12$, whereas Pave-GRPO attains the best HPS-v3 score ($14.62$) with only $+6$s overhead.
Meanwhile, it can also be observed from Tab.~\ref{tab:ablations} that every decomposition set $\mathcal{S}$ surpasses the baseline, confirming the robustness of the multi-path design; however, finer decompositions increase the wall-clock time with diminishing returns, so $\mathcal{S}=\{2,3\}$ is adopted throughout for its favorable efficiency--quality trade-off.

\begin{figure}[t]
  \centering
  \includegraphics[width=\linewidth]{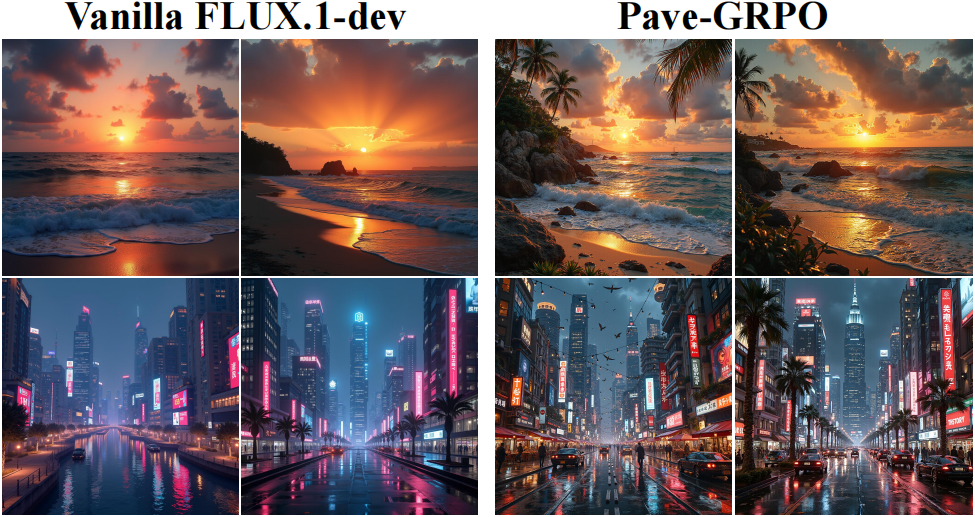}
  \caption{Diminished diversity after GRPO post-training. Prompts: ``A beautiful sunset over the ocean.'' (top) and ``The street scene on City Boulevard.'' (bottom).}
  \label{fig:limitation}
\end{figure}

\subsection{Limitation}
While improving preference alignment, Pave-GRPO suffers from quality--diversity trade-off. As shown in Fig.~\ref{fig:limitation}, the vanilla FLUX.1-dev generates markedly different compositions across seeds (e.g., open seascape versus rocky shoreline for the same sunset prompt), whereas after post-training the outputs collapse into nearly identical layouts. This stems from reward hacking: relative advantage estimation drives the policy toward high-reward trajectories, progressively reducing entropy and collapsing the sampling distribution onto a few high-reward modes.

\section{Conclusion}
In this work, we present Pave-GRPO, a framework that enhances GRPO post-training for flow model preference alignment without resorting to expensive high-step rollouts. It retains lightweight few-step pure-SDE sampling and decomposes each coarse trajectory into consecutive segments, leveraging the Markov property of iterative denoising and the shared Gaussian probability paths of ODE and SDE to convert cheap rollout advantages into dense supervision signals. This decouples generation budget from optimization granularity, extending the effective optimization range while implicitly enforcing piecewise velocity consistency through shared intermediate targets. The model thereby achieves more comprehensive preference alignment and stronger cross-metric generalization without additional sampling cost. Experiments on FLUX.1-dev and SD3.5-M confirm that this trajectory decomposition outperforms competing methods across diverse reward settings.

\bibliography{aaai2027}

% Check whether the conference requires a reproducibility checklist to be included in the paper.
% If so, you can uncomment the following line and ajust the path to include it.
% \input{ReproducibilityChecklist.tex}
\clearpage

\section{Pseudo code}
To facilitate understanding and reproducibility, we provide the detailed pseudo-code of Pave-GRPO in Algorithm~\ref{alg:pave-grpo}, which instantiates the three-stage pipeline described in the main text. Concretely, each training iteration consists of: (i) \emph{SDE group generation}, where a group of $G$ trajectories is rolled out from the current policy $\pi_{\theta_{\mathrm{old}}}$ via few-step pure-SDE sampling; (ii) \emph{advantage estimation}, where the terminal outputs are scored by the reward model and group-normalized into relative advantages; and (iii) \emph{multi-path optimization}, where each coarse transition is decomposed into hybrid ODE--SDE sub-trajectories of granularity $k \in \mathcal{S}$, and the resulting closed-form Gaussian kernels are evaluated at the same sampled endpoint to supervise the $k$ intermediate velocity predictions. The pseudo-code also makes explicit that the decomposition requires no additional rollout generation or reward evaluation: the only extra computation is the $k{-}1$ deterministic ODE sub-steps used to form the hybrid kernel means during the policy update. Equation numbers cited in Algorithm~\ref{alg:pave-grpo} refer to the main text.

\section{More examples}
\begin{figure*}[p]
    \centering
    \includegraphics[width=1.0\linewidth,height=0.9\textheight,keepaspectratio]{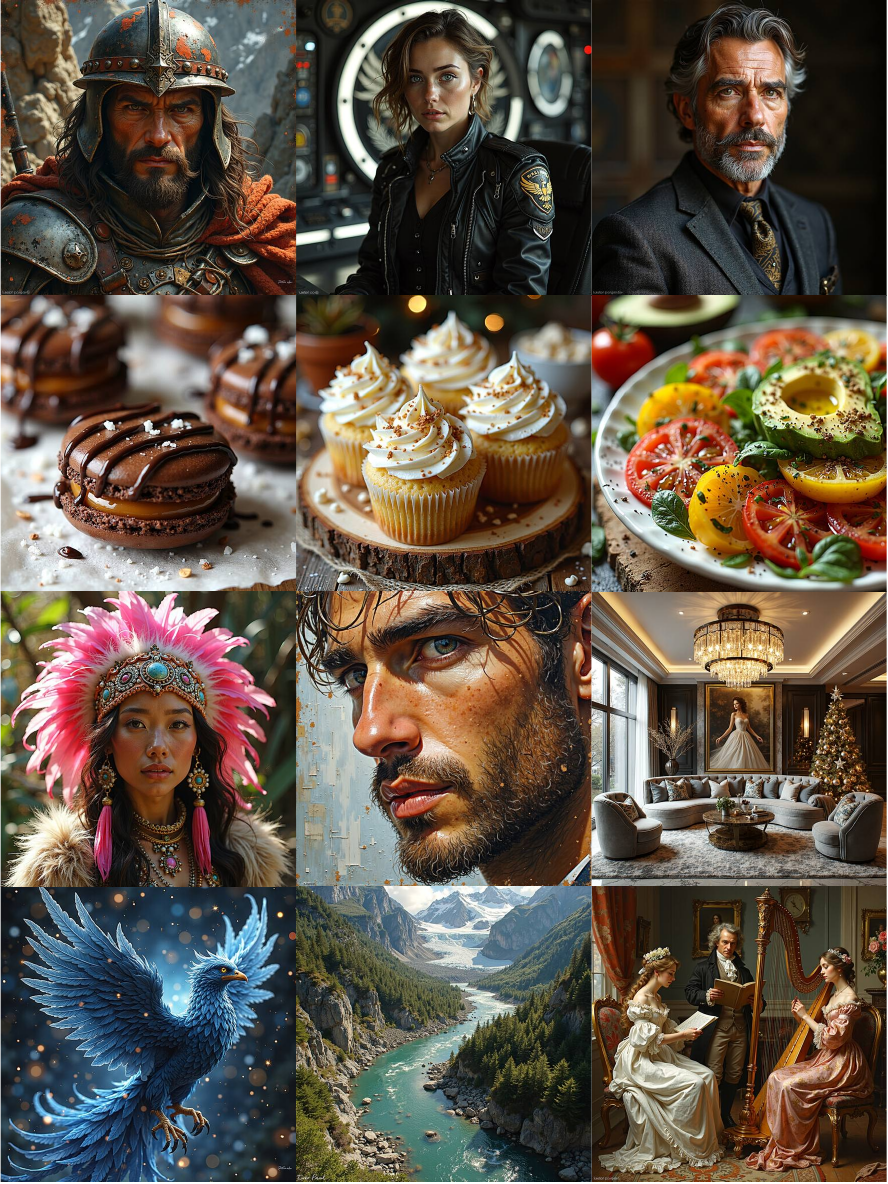}
    \caption{\textbf{More visual results of Pave-GRPO (1/4).}}
    \label{fig:more_vis_1}
\end{figure*}

\begin{figure*}[p]
    \centering
    \includegraphics[width=1.0\linewidth,height=0.9\textheight,keepaspectratio]{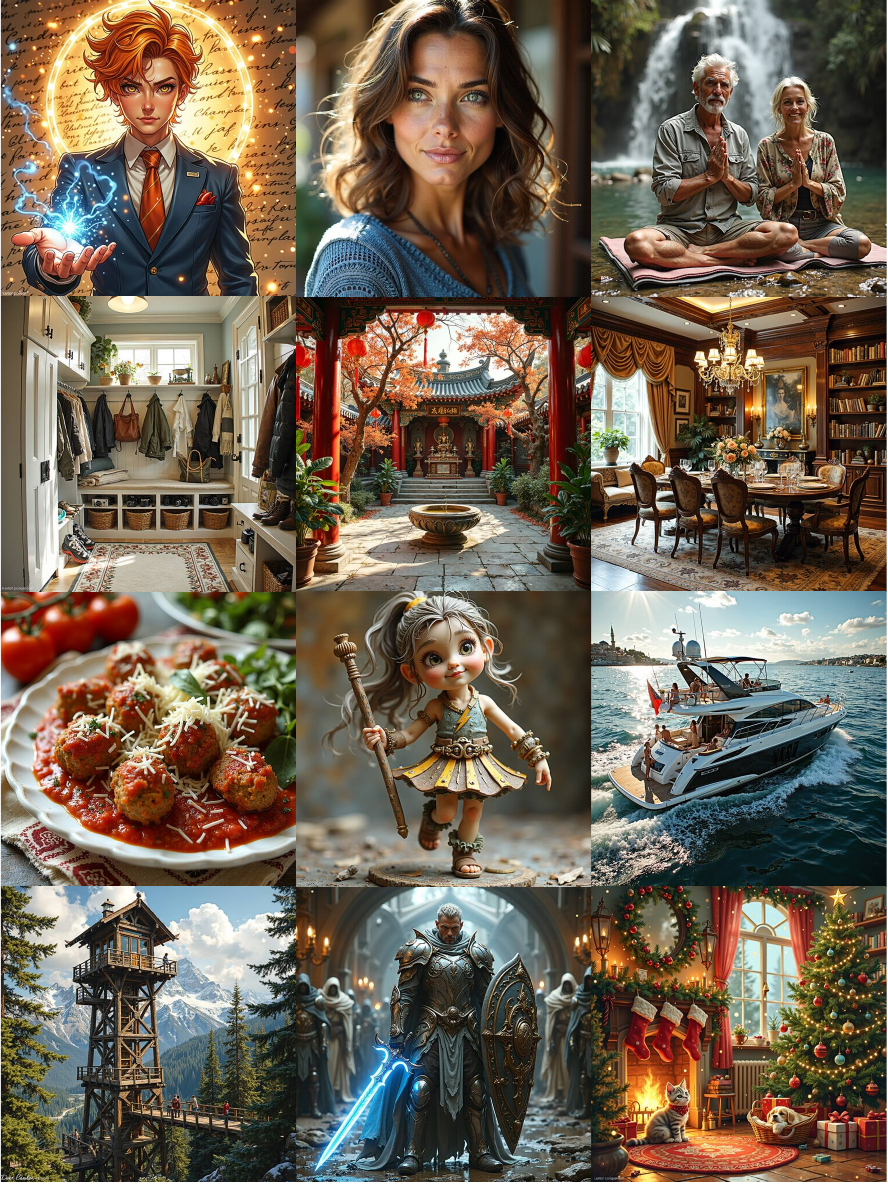}
    \caption{\textbf{More visual results of Pave-GRPO (2/4).}}
    \label{fig:more_vis_2}
\end{figure*}

\begin{figure*}[p]
    \centering
    \includegraphics[width=1.0\linewidth,height=0.9\textheight,keepaspectratio]{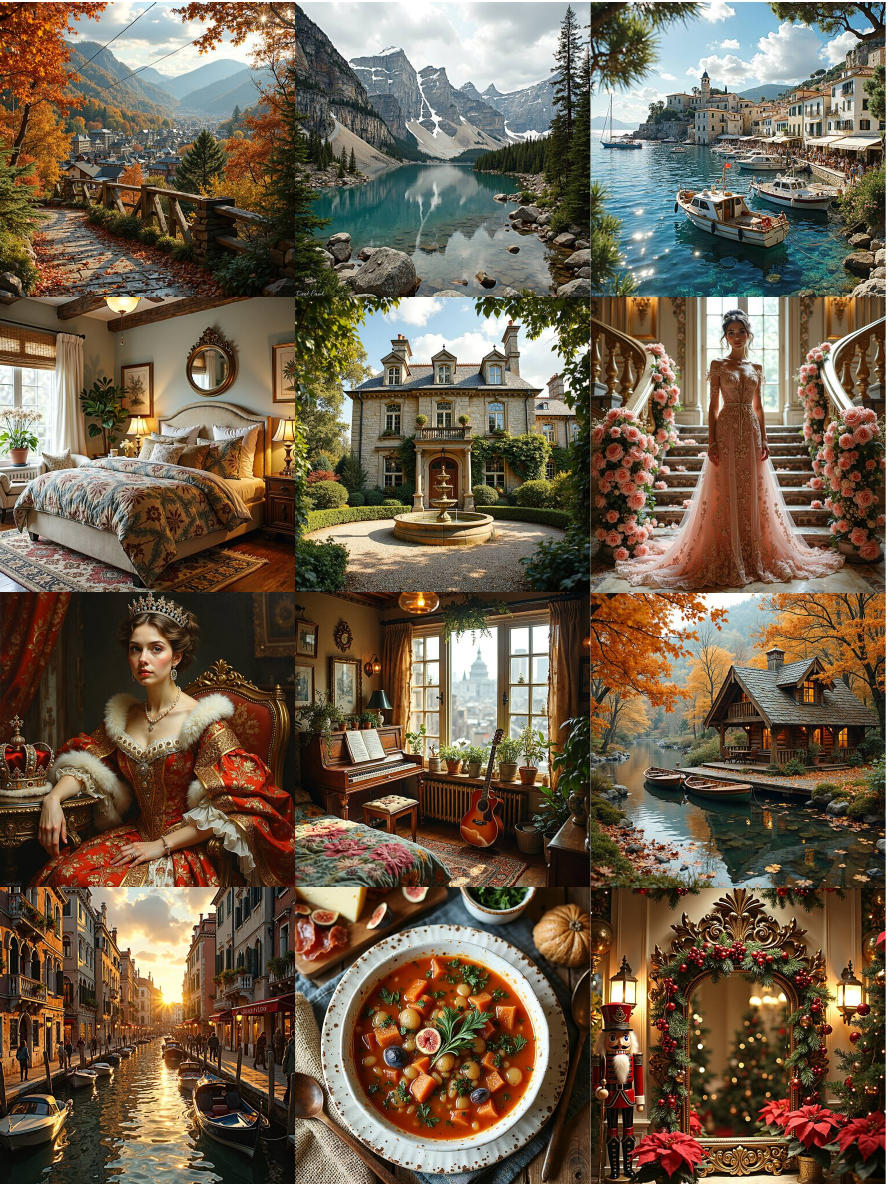}
    \caption{\textbf{More visual results of Pave-GRPO (3/4).}}
    \label{fig:more_vis_3}
\end{figure*}

\begin{figure*}[p]
    \centering
    \includegraphics[width=1.0\linewidth,height=0.9\textheight,keepaspectratio]{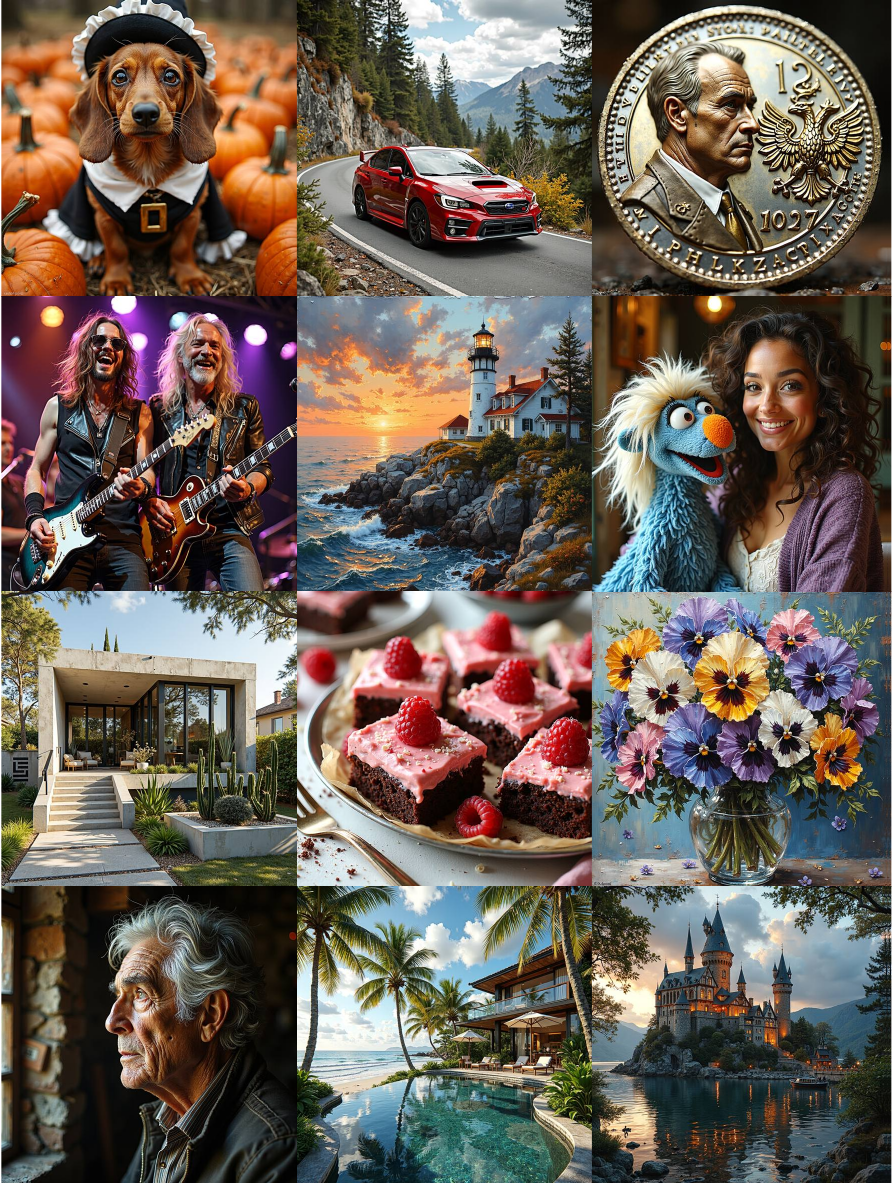}
    \caption{\textbf{More visual results of Pave-GRPO (4/4).}}
    \label{fig:more_vis_4}
\end{figure*}

To further demonstrate the effectiveness of Pave-GRPO, we present additional qualitative comparisons across diverse scenarios in Figs.~\ref{fig:more_vis_1}--\ref{fig:more_vis_4}. These examples cover a broad range of challenging generation conditions, including complex multi-object layouts, fine-grained attribute binding, stylized and photorealistic rendering, and difficult lighting and texture details.   These results corroborate the quantitative findings in the main text and further validate the robustness and generalization of the proposed trajectory decomposition across prompts and reward models. Please refer to these figures for detailed visual comparisons.

\section{Text prompts}
\label{sec:prompt}
For completeness and reproducibility, we provide in Tabs.~\ref{tab:prompt_main1}--~\ref{tab:prompt_supp4} the full set of text prompts used for image generation throughout this work, covering all qualitative results in the main text and this supplementary material. The prompts are organized in the order of the corresponding figures and span diverse content categories, including natural scenes, human portraits, stylized and anime renderings, object compositions, and scenes involving complex spatial layouts or fine-grained details, thus offering broad coverage of practical text-to-image scenarios. Unless otherwise specified, samples compared within the same case are generated with the same random seed under the default inference configuration, so that the visual differences are solely attributable to the compared methods. 

\begin{algorithm}[ht]
\caption{Pave-GRPO (one training iteration)}
\label{alg:pave-grpo}
\small
\begin{tabbing}
x\=x\=x\=x\=x\=x\=x\=x\kill
\textbf{Input:} policy $\pi_\theta$ ($\pi_{\mathrm{ref}} \leftarrow \pi_\theta$), reward model $R$, \\
\> prompt set $\mathcal{D}$, decomposition factors $\mathcal{S}$, group size $G$, \\
\> denoising steps $T$, clip range $\varepsilon$, KL coefficient $\beta$ \\
\textbf{for} each iteration \textbf{do} \\
\> $\theta_{\mathrm{old}} \leftarrow \theta$; \quad sample prompt $c \sim \mathcal{D}$ \\
\> \textbf{for} $i = 1$ \textbf{to} $G$ \textbf{do} \\
\> \> Sample $x_T^i \sim \mathcal{N}(0, I)$ \\
\> \> Rollout $\Gamma^i = (x_T^i, \dots, x_0^i)$ via $T$-step pure-SDE \\
\> \> \quad sampling with $\pi_{\theta_{\mathrm{old}}}$ \hfill (Eqs.~2--3) \\
\> \textbf{end for} \\
\> Rewards $r_i = R(x_0^i, c)$ \hfill (Eq.~4) \\
\> Advantages $\hat{A}^i = \bigl(r_i - \mathrm{mean}(\{r_j\})\bigr) \big/ \mathrm{std}(\{r_j\})$ \\
\> $\mathcal{J} \leftarrow 0$ \\
\> \textbf{for} $i = 1$ \textbf{to} $G$, \; $t = 1$ \textbf{to} $T$ \textbf{do} \\
\> \> $\rho \leftarrow p_\theta(x_{t-\Delta t}^i \,|\, x_t^i) \,/\, p_{\theta_{\mathrm{old}}}(x_{t-\Delta t}^i \,|\, x_t^i)$ \;// pure-SDE \\
\> \> $\ell \leftarrow \min\bigl(\rho \hat{A}^i, \, \mathrm{clip}(\rho, 1-\varepsilon, 1+\varepsilon) \hat{A}^i\bigr)$ \\
\> \> \textbf{for} $k \in \mathcal{S}$ \textbf{do} \\
\> \> \> Form $\mu^{(k)}, \Sigma^{(k)}$ via $k{-}1$ ODE sub-steps from $x_t^i$ \hfill (Eq.~9) \\
\> \> \> $\eta \leftarrow p_\theta^{(k)}(x_{t-\Delta t}^i \,|\, x_t^i) \,/\, p_{\theta_{\mathrm{old}}}^{(k)}(x_{t-\Delta t}^i \,|\, x_t^i)$ \;// Eq.~10 \\
\> \> \> $\ell \leftarrow \ell + \frac{1}{|\mathcal{S}|} \min\bigl(\eta \hat{A}^i, \, \mathrm{clip}(\eta, 1-\varepsilon, 1+\varepsilon) \hat{A}^i\bigr)$ \\
\> \> \textbf{end for} \\
\> \> $\mathcal{J} \leftarrow \mathcal{J} + \ell$ \\
\> \textbf{end for} \\
\> $\mathcal{J} \leftarrow \mathcal{J}/(GT) - \beta \, \mathcal{D}_{\mathrm{KL}}(\pi_\theta \,\|\, \pi_{\mathrm{ref}})$ \\
\> Update $\theta$ via gradient ascent on $\mathcal{J}$ \\
\textbf{end for}
\end{tabbing}
\end{algorithm}

\clearpage
\clearpage

\clearpage

\clearpage

\begin{table*}[!t]
    \centering
    \small
    \setlength{\tabcolsep}{4pt}
    \setlength{\aboverulesep}{0.5pt}
    \setlength{\belowrulesep}{0.5pt}
    \begin{tabular}{>{\arraybackslash}p{0.95\textwidth}}
        \toprule
        {London townhouse entrance with topiary and flowers.} \\
        \midrule
        A vibrant watercolor painting of a harbor scene featuring a red and white fishing boat docked at a pier. The boat is surrounded by other vessels, with masts and rigging visible. The water is depicted with blue and green washes, and the sky is a light gray. The painting has a loose, sketchy style with visible brushstrokes. Watercolor on paper, pen sketch. \\
        \midrule
        A stonechat perched on a thorny bush against a clear blue sky, with detailed feathers and a naturalistic setting. \\
        \midrule
        A cozy, rustic living room with a large fireplace, plush seating, and a stunning mountain view through floor-to-ceiling windows. The room features a chandelier, wooden beams, and a warm, inviting ambiance with a lit fireplace and candles on the coffee table. \\
        \midrule
        A man with long, dark hair and a beard is smoking a cigarette, with smoke rising from the tip. He is wearing a casual, worn-out jacket and has a can of soda in his other hand. The background is dark and smoky, emphasizing the cigarette smoke. \\
        \midrule
        Red cabin in snowy mountainscape. \\
        \midrule
        A fiery, demonic creature with glowing red eyes and a flaming mane stands menacingly amidst a fiery landscape, holding a large, ornate hammer with a glowing orange hue, surrounded by molten lava and rocky terrain. intense firelight. frontal view. ominous and fiery. \\
        \midrule
        Vibrant sunset over a swamp with reflection and distant hills. \\
        \midrule
        A top-down view of a ceramic cappuccino cup with ornate golden rim, resting on a matching saucer atop a rustic walnut wooden table. The latte art foam features an elaborate heart-shaped rosetta design with the stylized text Pave-GRPO elegantly inscribed in rich cinnamon-colored cocoa powder, accented by a small golden heart and delicate leaf flourishes. Warm morning sunlight streams in from the side, casting soft shadows and highlighting the velvety microfoam texture. Scattered around the cup are a few glossy coffee beans, dried orange berries, and a small glass jar of honey catching the light. A folded linen napkin and fresh green leaves peek into the frame. Photorealistic, cozy café aesthetic, shallow depth of field, 8K detail, food photography. \\
        \midrule
        A close-up of a person's face with delicate flowers partially covering the eyes and hair framing the face, creating a soft and ethereal effect. \\
        \midrule
        A variety of sushi, including nigiri and maki rolls, is arranged on a white ceramic plate with metal chopsticks placed beside it. The sushi is accompanied by small black and white dipping sauces. The setting includes a light stone background with a green leafy plant in the corner, creating a fresh and modern aesthetic. Soft and even, highlighting the sushi and creating a clean, bright appearance. Top-down view with a focus on the sushi and utensils, set against a light stone background. \\
        \midrule
        A freshly baked, golden-brown tart with a flaky crust and a generous topping of toasted pine nuts, dusted with powdered sugar, sits on a sheet of parchment paper. A single slice has been cut and placed in front of the tart, garnished with a sprig of rosemary. A vintage knife rests beside the tart on the parchment paper. Top-down view with a focus on the tart and its details. Even and bright, highlighting the textures and colors of the tart. Warm and inviting, offering a comforting and delicious treat. \\
        \midrule
        A serene landscape featuring Bridal Veil Falls and the Merced River in Yosemite Valley, with towering granite cliffs bathed in warm, golden light at sunset, reflecting off the river's surface, surrounded by dense evergreen forests and a fallen tree partially submerged in the water. Golden sunset light. Peaceful and majestic. \\
        \midrule
        Anime character with spiky hair and glowing aura. anime. glowing blue aura. front view. digital art. \\
        \midrule
        A woman in a Renaissance-style dress sits at a table adorned with a bouquet of flowers, a gold chalice, and various jewelry and coins, symbolizing the transient nature of wealth and beauty. Oil on canvas. Frontal view, seated. Soft, even lighting. Renaissance. Serene, contemplative. \\
        \midrule
        A close-up of a man with a mustache and short hair, wearing a dark green shirt, looking intently at something off-camera with a serious expression. \\
        \midrule
        A luxurious balcony with a dining table set for four, featuring green plates and silverware, surrounded by wicker chairs and ottoman. The backdrop showcases a stunning view of the ocean and lush greenery, with a ceiling fan and large sliding glass doors enhancing the serene atmosphere. \\
        \midrule
        A helicopter with vibrant red and blue markings flies over a majestic waterfall, with mist and a rainbow visible in the spray. The helicopter is positioned slightly to the left of the waterfall, and the surrounding landscape features lush greenery and a river with rapids. Dynamic and adventurous. Aerial view. Natural daylight. \\
        \midrule
        A portrait of a spiritual warrior adorned with a crown of leaves and red berries, holding a xoloitzcuintle dog and an incense burner, surrounded by marigold flowers and candles, with the inscription 'Vidi Vici' on the warrior's chest. even. mystical and serene. painting. oil on canvas. \\
        \bottomrule
    \end{tabular}
    \caption{Generation prompts for Fig.~1 of the main text, listed in the order from left to right and top to bottom.}
    \label{tab:prompt_main1}
\end{table*}

\clearpage

\begin{table*}[!t]
    \centering
    \small
    \setlength{\tabcolsep}{4pt}
    \setlength{\aboverulesep}{0.5pt}
    \setlength{\belowrulesep}{0.5pt}
    \begin{tabular}{>{\arraybackslash}p{0.95\textwidth}}
        \toprule
        \multicolumn{1}{l}{\textbf{Prompts for Fig.~3 (Main.)}} \\
        \midrule
        A photo of a puppy playing on the lawn. \\
        \midrule
        \multicolumn{1}{l}{\textbf{Prompts for Fig.~4 (Main.)}} \\
        \midrule
        {A 3D rendering of anime schoolgirls with a sad expression underwater, surrounded by dramatic lighting.} \\
        \midrule
        A happy daffodil with big eyes, multiple leaf arms and vine legs, rendered in 3D Pixar style. \\
        \midrule
        A cute little anthropomorphic Tropical fish knight wearing a cape and a crown in short, pale blue armor. \\
        \midrule
        A female human barbarian depicted in a traditional Dungeons and Dragons illustration. \\
        \midrule
        A fruit basket on a kitchen table with a Studio Ghibli reference. \\
        \midrule
        A train that is going by a building \\
        \bottomrule
    \end{tabular}
    \caption{Generation prompts for Fig.~3 and Fig.~4 of the main text, grouped by figure and listed in the order from left to right and top to bottom.}
    \label{tab:prompt_main34}
\end{table*}

\begin{table*}[!t]
    \centering
    \small
    \setlength{\tabcolsep}{4pt}
    \setlength{\aboverulesep}{0.5pt}
    \setlength{\belowrulesep}{0.5pt}
    \begin{tabular}{>{\arraybackslash}p{0.95\textwidth}}
        \toprule
        A digital painting of a rugged, battle-worn individual with a stern expression, wearing a helmet with a red and black visor, set against a dark, textured background with a mountainous landscape in the distance. The character has long, dark hair and a beard, and the overall tone is gritty and intense. Digital art, created using a painting software with a focus on texture and shading. Digital painting. Close-up of the character's face and upper body, with the background slightly blurred to emphasize the subject. Subtle, with shadows highlighting the character's features and the texture of the background. \\
        \midrule
        A woman with short brown hair and blue eyes is seated in a dark, high-tech environment, wearing a black zip-up jacket. She is in a serious or contemplative mood, and the background features a circular emblem with a bird-like symbol, indicating a military or scientific setting. \\
        \midrule
        Serious man in blazer against dark background. \\
        \midrule
        Close-up side view of several dark chocolate caramel dairy-free macarons with a glossy caramel center, drizzled with dark chocolate sauce and sprinkled with sea salt on a white parchment paper background. Side view with focus on a single macaron in the foreground, slightly blurred macarons in the background. \\
        \midrule
        Vanilla cupcakes with white frosting and sprinkles on a wooden stand. \\
        \midrule
        A close-up of a fresh salad featuring vibrant slices of red and yellow tomatoes, creamy avocado slices, and a drizzle of olive oil with herbs and spices sprinkled on top, presented on a white plate. \\
        \midrule
        A person wearing a vibrant headdress adorned with bright pink feathers, intricate beadwork, and fur, set against a natural outdoor backdrop with greenery and sunlight filtering through the leaves. \\
        \midrule
        Impressionistic close-up portrait of a man's face. Close-up. Impressionistic. Oil paint. \\
        \midrule
        A luxurious living room with a large, modern chandelier featuring multiple glass and brass elements hanging from the ceiling. The room is furnished with a curved gray sofa adorned with patterned cushions, two matching armchairs, and a unique, textured coffee table. A small, decorated Christmas tree stands to the right, and a large artwork of a woman in a flowing dress hangs on the wall behind the sofa. The space is bright and airy with large windows allowing natural light to flood in. Digital rendering. Elegant and sophisticated. \\
        \midrule
        A stylized blue phoenix with intricate feather patterns, wings spread wide, soaring against a dark, starry night sky with glowing particles trailing behind it. mystical, ethereal. digital illustration. glowing particles, dark background. digital art. side view, dynamic pose. \\
        \midrule
        An aerial view of a rugged, mountainous landscape featuring a large, winding river cutting through a valley, with a glacier in the background and a small body of water near the river's mouth. Aerial. \\
        \midrule
        Elegant figures making music in an interior. Neoclassical. The figures are arranged in a semi-circle, with the woman reading and the woman playing the harp in the foreground, and the man with sheet music in the background. The composition is balanced with the large harp as the focal point. Soft, even lighting typical of indoor paintings from the 18th or 19th century, with no harsh shadows. Oil on canvas. Serene and refined. \\
        \bottomrule
    \end{tabular}
    \caption{Generation prompts for Fig.~\ref{fig:more_vis_1} of this supplementary material, listed in the order from left to right and top to bottom.}
    \label{tab:prompt_supp1}
\end{table*}

\begin{table*}[!t]
    \centering
    \small
    \setlength{\tabcolsep}{4pt}
    \setlength{\aboverulesep}{0.5pt}
    \setlength{\belowrulesep}{0.5pt}
    \begin{tabular}{>{\arraybackslash}p{0.95\textwidth}}
        \toprule
        A vibrant anime character with fiery orange hair and a determined expression, dressed in a sharp blue suit, holding out their hand with a glowing, ethereal energy emanating from it, set against a backdrop of swirling, handwritten script and a radiant light source. frontal view, character centered. digital illustration. anime. bright, radiant, ethereal. determined, powerful. \\
        \midrule
        Close-up portrait of a woman with shoulder-length brown hair, wearing a blue top, looking slightly to the side with a thoughtful expression, natural lighting, and a soft focus background. \\
        \midrule
        Middle-aged couple with yoga mats in front of a waterfall. \\
        \midrule
        A spacious, well-organized mudroom with white cabinetry, multiple coat hooks, storage baskets, and shelves holding various items like jackets, umbrellas, and cameras. The floor is tiled with a decorative border, and the space is bright and airy with natural light coming from a window above. \\
        \midrule
        A traditional Chinese temple courtyard adorned with vibrant red prayer ribbons hanging from trees, leading to a central altar with statues and offerings, surrounded by potted plants and a stone basin. \\
        \midrule
        Formal dining room with bookshelf and chandelier. \\
        \midrule
        A plate of meatballs in a rich tomato sauce, topped with shredded cheese, served with fresh tomatoes and a side salad on a patterned tablecloth. \\
        \midrule
        A clay sculpture of a young girl in mid-dance, wearing a skirt with a lightning bolt design and a belt, holding a stick in her right hand, her left leg lifted high, and her hair tied back in a ponytail. \\
        \midrule
        A luxurious motor yacht with a sleek white and black exterior is cruising on calm blue waters. Several people are enjoying the deck, some standing and others sitting, while a person is at the helm. The yacht features a spacious open deck with wooden flooring and a flag at the stern. The background shows a distant shoreline with buildings and a clear blue sky. Bright and natural daylight with clear visibility. Relaxed and leisurely. Aerial view capturing the yacht and its surroundings. \\
        \midrule
        A towering wooden observation tower rises high above a dense forest, with visitors walking along its elevated walkways. The structure is supported by sturdy beams and stands against a backdrop of snow-capped mountains and a clear sky. The scene captures the essence of a serene and adventurous outdoor experience. glowing particles, dark background. digital art. side view, dynamic pose. \\
        \midrule
        A powerful warrior in futuristic armor stands in a dimly lit, post-apocalyptic environment. The character wields a glowing blue sword and shield, with a glowing blue light emanating from the sword's hilt. The background features ghostly figures and a sense of impending battle, with a dark, ominous atmosphere. Digital illustration. Digital art. Mysterious and intense, with a sense of anticipation and danger. Centralized figure with a focus on the warrior and his glowing sword, with a dark, atmospheric background. \\
        \midrule
        A cozy Christmas scene with a decorated Christmas tree, a fireplace with stockings, a cat playing with a toy, and a dog sleeping in a basket. The room is filled with festive decorations and presents under the tree. Cartoon. Digital illustration. Warm. Interior. \\
        \bottomrule
    \end{tabular}
    \caption{Generation prompts for Fig.~\ref{fig:more_vis_2} of this supplementary material, listed in the order from left to right and top to bottom.}
    \label{tab:prompt_supp2}
\end{table*}

\begin{table*}[!t]
    \centering
    \small
    \setlength{\tabcolsep}{4pt}
    \setlength{\aboverulesep}{0.5pt}
    \setlength{\belowrulesep}{0.5pt}
    \begin{tabular}{>{\arraybackslash}p{0.95\textwidth}}
        \toprule
        Gatlinburg, Tennessee, autumn foliage and mountains. Natural daylight. Panoramic view. Serene and picturesque. \\
        \midrule
        Reflective lake with rugged mountains and misty sky. A wide-angle shot capturing the expansive view of the lake, mountains, and sky. Soft, diffused lighting without harsh shadows, indicating an overcast day. Peaceful and awe-inspiring. \\
        \midrule
        Island harbor with fishing boat and white buildings. View from the sea, looking towards the island. Bright and sunny, with clear shadows indicating midday sunlight. \\
        \midrule
        Cozy bedroom with queen-sized bed, beige headboard, and geometric comforter. \\
        \midrule
        A grand historic stone building with arched windows and a slate roof, surrounded by a well-maintained garden featuring a central fountain and neatly trimmed hedges. The facade boasts a central entrance with double doors and a balcony above, bathed in natural light, indicating a clear day. Natural daylight. \\
        \midrule
        A woman stands on a grand staircase, dressed in an elegant floor-length blush pink gown adorned with floral patterns, surrounded by cascading pink roses. The opulent setting features ornate railings and a large window, indicating a luxurious indoor venue. Soft, even lighting highlighting the subject and the intricate details of the gown and floral decorations. \\
        \midrule
        A regal portrait of a woman in a luxurious red and gold gown, adorned with intricate embroidery and a fur-trimmed cape, seated on an ornate chair with a crown resting on a table beside her. Her posture is poised and dignified, with her hand resting on the table, and she wears a pearl necklace and earrings. The background is dark and simple, focusing attention on the subject. Frontal view, seated, with crown on table. Soft, even lighting highlighting the subject and her attire. Oil painting. Formal and dignified. \\
        \midrule
        A cozy bedroom with a wooden piano and a guitar, a large window with a cityscape view, and a bed with a colorful floral bedspread. \\
        \midrule
        A serene autumn scene featuring a wooden boathouse with two boats docked under its roof, surrounded by vibrant fall foliage and a calm river reflecting the colors of the trees. The setting is peaceful and picturesque, evoking a sense of tranquility and nostalgia. \\
        \midrule
        Venetian canal at sunset. tranquil. photograph. golden sunset. photographic. \\
        \midrule
        A rustic autumnal minestrone soup served in a white bowl, accompanied by a wooden spoon, a wooden cutting board with cheese, prosciutto, figs, and parsley, and a breadstick on a burlap and denim placemat. Top-down view. \\
        \midrule
        A luxurious Christmas display featuring an ornate wooden mirror with intricate carvings, adorned with a garland of red berries and greenery, surrounded by festive decorations including a nutcracker and poinsettias. Photograph. Bright and evenly lit. Elegant and festive. \\
        \bottomrule
    \end{tabular}
    \caption{Generation prompts for Fig.~\ref{fig:more_vis_3} of this supplementary material, listed in the order from left to right and top to bottom.}
    \label{tab:prompt_supp3}
\end{table*}

\begin{table*}[!t]
    \centering
    \small
    \setlength{\tabcolsep}{4pt}
    \setlength{\aboverulesep}{0.5pt}
    \setlength{\belowrulesep}{0.5pt}
    \begin{tabular}{>{\arraybackslash}p{0.95\textwidth}}
        \toprule
        A dachshund dog dressed in a pilgrim costume stands amidst a pile of pumpkins, looking directly at the camera with a curious expression. \\
        \midrule
        A red Subaru WRX driving on a winding mountain road with a rocky cliff and evergreen trees in the background under a partly cloudy sky. \\
        \midrule
        A Polish commemorative coin featuring a portrait of Witold Pilecki and the Polish eagle emblem, with the inscription '10 zł' and '2017' on the reverse side. \\
        \midrule
        Two musicians performing on stage, one playing a guitar and the other a bass, both wearing leather and casual attire, with vibrant stage lighting in the background. Energetic and lively. Vibrant stage lighting with spotlights and purple hues. \\
        \midrule
        Lighthouse on rocky cliff at dusk. Oil painting on canvas. Peaceful and contemplative. Impressionistic. \\
        \midrule
        A woman and a blue puppet with a fluffy white mane are smiling at each other in a warm indoor setting. The woman has dark curly hair and wears a purple cardigan over a white top. The puppet has a friendly expression with large round eyes and an orange beak. The background features a beige wall and a green door, creating a cozy home-like environment. Warm and friendly. \\
        \midrule
        A modern house with a minimalist design, featuring a large window with black frames and a concrete facade. The house has a staircase leading up to the entrance, and there is a large planter box with cacti in front of it. The sky is clear and blue, indicating a sunny day. \\
        \midrule
        A close-up of several square pieces of fudgy dark chocolate brownies topped with a smooth, pink sugar-free Greek yogurt raspberry frosting, garnished with fresh raspberries, arranged neatly on a round metal tray with parchment paper underneath, and a fork placed nearby on a white surface. Bright, even lighting with no visible shadows. \\
        \midrule
        A vibrant, abstract painting of a bouquet of pansies in a vase, with bold brushstrokes and a rich palette of blues, purples, pinks, and greens, creating a lively and dynamic floral arrangement. Impressionistic. Close-up of a bouquet in a vase. Oil on canvas. Lively and dynamic. \\
        \midrule
        A close-up side profile of an elderly man with gray hair, wearing a dark jacket and a striped shirt, gazing thoughtfully to the side in a dimly lit, rustic setting. Serene and contemplative. Soft, natural light highlighting the subject's face and hair, with a dark, shadowy background. \\
        \midrule
        A luxurious beachfront villa with an infinity pool, surrounded by tall palm trees and overlooking the ocean under a bright blue sky with scattered clouds. serene and luxurious. photograph. photorealistic. wide-angle view. bright and natural. \\
        \midrule
        A majestic medieval castle perched atop a rocky island, surrounded by calm waters reflecting its illuminated facade at dusk, with a serene sky transitioning from blue to orange hues. Warm, ambient lighting from the castle's lights reflecting on the water. Serene and majestic. \\
        \bottomrule
    \end{tabular}
    \caption{Generation prompts for Fig.~\ref{fig:more_vis_4} of this supplementary material, listed in the order from left to right and top to bottom.}
    \label{tab:prompt_supp4}
\end{table*}

\end{document}